\documentclass[10pt,a4paper]{article}

\usepackage{cogsci}

\cogscifinalcopy % Uncomment this line for the final submission 
\usepackage{graphicx}

\usepackage{pslatex}

\usepackage[bookmarks=false,hidelinks]{hyperref}
\usepackage{apacite}
\usepackage{float} % Roger Levy added this and changed figure/table
\usepackage{tabularx}
\usepackage{booktabs}
\usepackage{multirow}
\usepackage{amsmath}

\newcolumntype{P}[1]{>{\centering\arraybackslash}p{#1}}
\newcolumntype{R}{>{\raggedleft\arraybackslash}X}
\newcolumntype{C}{>{\centering\arraybackslash}X}

\title{Latent Event-Predictive Encodings through Counterfactual Regularization}

\author{
{\large \bf Dania Humaidan (dania.humaidan@uni-tuebingen.de)}$^{1,2}$ \\
{\large \bf Sebastian Otte (sebastian.otte@uni-tuebingen.de)}$^{1,2}$\\
{\large \bf Christian Gumbsch (christian.gumbsch@uni-tuebingen.de)}$^{1,2,3}$\\
{\large \bf Charley Wu (charley.wu@uni-tuebingen.de)}$^{2,4}$ \\
{\large \bf {Martin V. Butz (martin.butz@uni-tuebingen.de)} $^{1,2}$} \\
\vspace{0.0cm} \\	 
 $^1$  Neuro-Cognitive Modeling Group, Dep. of Psychology and Dep. of Computer Science, Sand 14, 72076 Tübingen, Germany \\
 $^2$ Cluster of Excellence – Machine Learning for Science, Maria-von-Linden-Str. 6, 72076, Tübingen, Germany\\
 $^3$: Autonomous Learning Group, MPI for Intelligent Systems, 72076, Tübingen, Germany\\
 $^4$ Human and Machine Cognition Lab, Maria-von-Linden-Str. 6,
 72076, Tübingen, Germany\\
}

\begin{document}

\maketitle

\begin{abstract} %150 word limit
A critical challenge for any intelligent system is to infer structure from continuous data streams.
Theories of event-predictive cognition suggest that the brain segments sensorimotor information into compact event encodings, which are used to anticipate and interpret environmental dynamics. 
Here, we introduce a SUrprise-GAted Recurrent neural network (SUGAR) using a novel form of counterfactual regularization. 
We test the model on a hierarchical sequence prediction task, where sequences are generated by alternating hidden graph structures.
Our model learns to both compress the temporal dynamics of the task into latent event-predictive encodings and anticipate event transitions at the right moments, given noisy hidden signals about them.
The addition of the counterfactual regularization term ensures fluid transitions from one latent code to the next, whereby the resulting latent codes exhibit compositional properties. 
The implemented mechanisms offer a host of useful applications in other domains, including hierarchical reasoning, planning, and decision making.

\end{abstract}

% \cite \citeA \citeNP
\section{Introduction}

Several disciplines have proposed that our brains construct generative, event-predictive models to interpret and shape our perception of the world \cite{Baldwin2021,Butz:2017,Butz:2021,franklin2020structured,Kuperberg2021,Zacks:2007}. 
Latent, event-predictive encodings provide top-down contextual guidance, which improves our prediction of local temporal patterns. 
As we watch scenes from a movie or experience daily life, our learned event models allow us to anticipate switches between events.
For instance, we can anticipate when the credits will roll or when the lecture is about to end, enabling us to suitably adjust our behavior in preparation of the transition.

%Our focus
Here we focus on the problem of learning to generate smooth event transitions and to thus enable the fluent processing of continuous streams of sensorimotor information across events. 
To achieve this, the learning of compact, event-predictive encoding seems to be of particular importance \cite{Baldwin2021,Zhu:2020}.
Accordingly, it has been suggested that our brains learn to preempt surprise, that is, to predict unpredictability when a current event is about to end \cite{Baldwin2021,franklin2020structured,Kuperberg2021}. 
Temporary increases in prediction error have been suggested to offer suitable event segmentation clues \cite{Zacks:2007,Zacks2011}.
Others have emphasized the importance of temporary stable latent predictive encodings between transitions \cite{Shin2021,schapiro2013neural}, which is generally compatible with surprise-biased processing.
Thus, while surprise marks event transitions, predictability marks ongoing events.

Here we present a SUrprise-GAted Recurrent neural network (SUGAR), as a biologically-inspired normative model that offers computational and algorithmic explanations for how compact, compositional event-predictive encodings may develop.
Similar to how uncertainty modulates the arbitration between systems in other domains \cite{lee2014neural, daw2005uncertainty}, we show that surprising sensory information can be used to modulate top-down control, thus fostering better encoding of latent events and transitions between them. 
However, once surprise is detected, only reactive or retrospective adaptations are possible. 
Thus, it is critical to preempt surprise when intending to process event transitions fluidly.

In the real world, orienting reflexes help us to investigate the causes of a surprising event, such as when we trip over some overlooked item.
As a result, we learn new causal relationships and refine the involved event-predictive encodings \cite{Zacks:2007}.
Moreover, we learn to preempt the surprise by preparing for an upcoming transition and thus switching fluently to the new events---as when we switch from walking to climbing up some stairs or when we open the fridge to get milk for our tea \cite{Baldwin2021,Kuperberg2021}. 
Accordingly, we augment SUGAR with an event boundary anticipation module, which is designed to learn to control event switches.
We show that as long as sufficient information is provided to anticipate event switches, the module learns to switch in time. 
However, fluctuations and unnecessary openings still occur. 
We show that the addition of a novel counterfactual regularization (CFR) loss can improve learning reliability as well as the precision and compactness of the developing latent event-characterizing encodings. CFR evaluates whether an event switch was indeed helpful for avoiding surprise, effectively balancing costs of event switches with rewards from preventing surprise.

We show that SUGAR with CFR is able to both anticipate event transitions and transition between latent event codes fluently, fully avoiding surprising outcomes.  
Moreover, we show that the emergent latent codes exhibit compositional properties of the hierarchical task dynamics. 
Our model architecture and learning techniques may be useful for both modeling cognition and improving the performance of artificial systems in a variety of reasoning, planning, and decision-making tasks.

\begin{figure}[t]
\includegraphics[width=0.98\columnwidth]{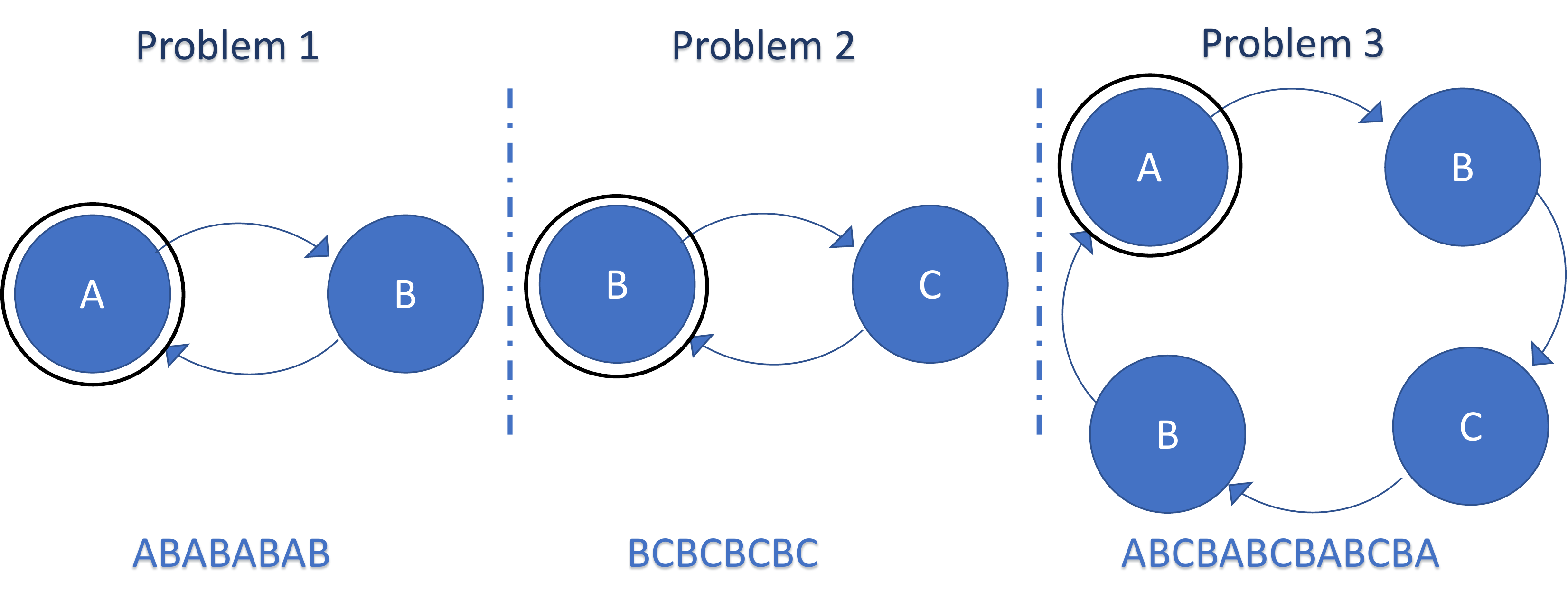}
\vspace{-.4cm}
\caption{Sequence of letters are generated from one of the three problems. Switching between the problems occurs at uniformly sampled time points within the interval of [10,30]. Each problem starts from the highlighted letter.} 
\label{comp_problem}
\end{figure}

% task description and model comes h
\section{Hierarchical Sequence Prediction Task}
In this paper we focus on the prediction of simple, hidden grammatical regularities within a sequence prediction task. 
The general structure of the task is shown in \autoref{comp_problem}.
Three events, or problems, are encoded as a systematic progression through a sequence of symbols: $[A,B]^\star$,
$[B,C]^\star$, and $[A,B,C,B]^\star$.
Transitions between the problems occur randomly, where the number of steps before the next transition $\tau$ is sampled uniformly randomly from $\tau \in \{10, \dots, 30\}$.
The sequences are fed sequentially into the SUGAR model as one-hot encoded sensory information $x^t$, which is provided to the event processing layer (cf. \autoref{models}). 
In the experiments we combine either problems 1 and 2, denoted by Problem 1+2, or all three, which we denote by Problem 1-3.

In addition to the symbol sequence, SUGAR's event anticipation layer (cf. \autoref{models}) receives additional information about which event is going to come next, denoted as contextual information (CI) input. 
Again by means of a one-hot enocding, CI indicates from a randomly chosen point in time $\tau^\star<\tau$ onward which problem will be sampled once the next event switch occurred.
Moreover, masked information is provided about the number of iterations $\tau$ before an event switch will occur, called the event boundary (EB) input. 
This EB input mimics information in the real world that helps us to anticipate that a new event is about to start. 
For example, we get ready to grasp a glass while reaching for it, anticipating the grasp initiation by tracking the distance between our hand and the glass.
In the experiments presented below, we first provide simple one-hot encoded switching information with additional randomly fluctuating inputs, challenging the learner to identify the correct switching signal. 
Later, we provide multiple relevant inputs in the form of an input vector whose activities tend to increase while the switch comes close. 
The actual switch occurs when all input vector values have reached full activity.

\section{The SUGAR model}
\begin{figure*}[t]
\includegraphics[width=\linewidth]{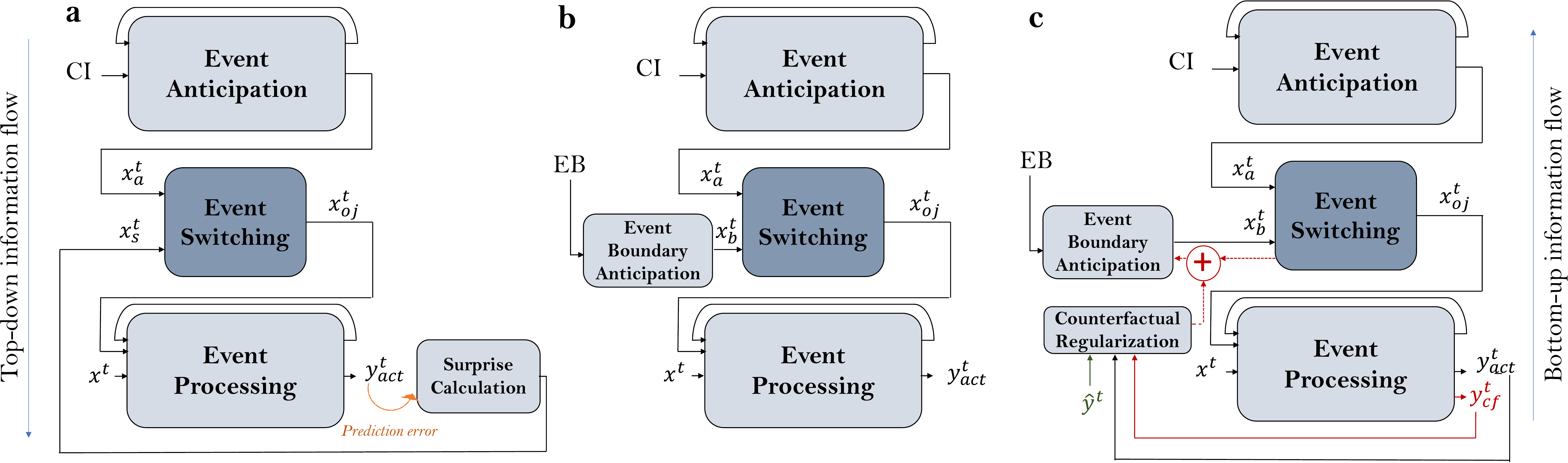}
\vspace{-.4cm}
\caption{\textbf{a}) SUGAR$_a$ calculates a surprise signal online and uses this signal to control the update gate in the event switching module.
\textbf{b}) SUGAR$_b$ replaces the surprise signal with a dedicated event boundary anticipation module, which receives fuzzy event boundary information. 
\textbf{c}) SUGAR$_c$ adds CFR, to foster more precise gate opening and thus the emergence of more compact latent event encodings. CI: contextual information input, EB: event boundary information input.} 
\label{models}
\end{figure*}

The SUGAR model includes four main event-predictive layers (see Fig.~\ref{models}). 
An \emph{event processing} layer is designed to process the currently unfolding event dynamics by predicting the next letter in the sequence.
An \emph{event anticipation} layer processes event-predictive embeddings to provide contextual guidance about which of the three problems (Fig.~\ref{comp_problem}) is currently active. This guidance is provided in the form of latent event encodings, which are sent to the event processing layer. 
Third, an \emph{event switching} layer acts as a gate controlling whether top-down contextual guidance is fed into the event processing layer. 
Finally, an \emph{event boundary} layer is designed to learn to predict when the next event transition will take place.

The event processing, anticipation, and boundary layers are implemented by long short-term RNNs \cite<LSTMs; >{hochreiter1997}, although other architectures may also be suitable these types of problem.
In our current problem, the event processing layer receives as input the sequence value $\mathbf{x}_{t-1}$ from the previous time step $t-1$, along with current top-down information $\mathbf{x}_{o}^{t}$ from the event switching layer. 

We investigate three progressively enhanced SUGAR variants in this paper, which control the gating mechanism in the event switching layer in distinct manners (Fig.~\ref{models}).
We first describe the exact processing dynamics of the switching layer.

\subsection{Event Switching Layer}
The event switching layer is implemented using a state switching gate, which are also used in gated recurrent unit RNNs \cite<GRU;>{cho2014}. The switching gate thus controls when the top-down guidance signal from the event anticipation layer will be passed down to the event processing layer.
Input to the event switching layer comes from the event anticipation layer ($\mathbf{x}_a$) and the event boundary layer ($\mathbf{x}_b$).
We denote its input layer with $\eta$ and its update gate with $\zeta$.

\subsubsection{Forward pass}

The scalar activity $x_{\zeta}$ of the update gate controls the switch between maintaining the previous latent event code or updating it with novel contextual information:
It is defined by:
\begin{align}
    net_{\zeta}^{t} &= x^t_s  \\
    x_{\zeta}^{t} &= \varphi_{\zeta} (net_{\zeta}^{t}), \nonumber 
\end{align}
where $x^t_s$ denotes the surprise signal input at the current point in time $t$, and $\varphi_{\zeta}$ denotes the cell's sigmoid activation function. 

The activity $\mathbf{x}_{\eta}$ of the input layer is determined by combining and integrating the output of the event anticipation layer $\mathbf{x}_{a}$ as well as the previous hidden state of the event switching layer $\mathbf{x}_{h}$:
\begin{align}
    net_{\eta j}^{t} &= \sum_i w^a_{ij} x_{ai}^{t} + \sum_{j'} w^{\eta}_{j'j} x_{hj'}^{t-1}  \\
    x_{\eta j}^{t} &= \varphi_{\eta} (net_{\eta j}^{t}), \nonumber
\end{align}
where the activation function $\varphi_{\eta}$ is linear, $i$ is the index of the input layer unit, $j$ and $j'$ are the indices of the hidden layer unit.

The hidden cell state $\mathbf{x}_{h}$ is then determined by fusing the input layer activities with the previous hidden layer activities, dependent on the update gate's current activity: 
\begin{equation}
    x_{hj}^{t} = x_{\zeta}^{t} x_{hj}^{t-1} + (1-x_{\zeta}^{t}) x_{\eta j}^t
\end{equation}

Finally, the output of the event switching layer, fed to the event processing layer, is defined using a regular feed-forward pass: 
\begin{align}
    net_{o j}^{t} &= \sum_i w^o_{ij} x_{hi}^t   \\
    x_{o j}^{t} &= \varphi_{o} (net_{o j}^{t}), \nonumber
\end{align}
where the respective activation function $\varphi_{o}$ is linear.

\subsubsection{Backwards pass}
To update the parameters of the network, standard back-propagation through time is used throughout the model, which adapts the weights using ADAM. 
The error signal defined by the squared reconstruction loss, which is determined at the prediction output of the event processing layer. Thus, the full model is trained end-to-end.

\subsection{Event Boundary Processing}

In this work, we consider three different variants of SUGAR, concerning how surprise is processed and thus how event boundaries are anticipated.
First, we calculate a surprise signal by monitoring the prediction errors in the event processing module (Fig.~\ref{models}a).
As expected, the resulting system responds to event boundaries after the fact, instead of anticipating future changes.
Second, we provide fuzzy, hidden ground truth information about event boundaries, which is processed by the event boundary anticipation module and then passed onto the event switching module via $x_s^t$ (Fig.~\ref{models}b). Lastly, we add counterfactual regularization (CFR) to improve event boundary predictions further (Fig.~\ref{models}c).

\subsubsection{Explicit Surprise Processing}
We first implemented SUGAR$_a$, which calculates surprise explicitly (Fig.~\ref{models}a) and invokes event transitions upon its detection. 
Surprise is calculated by maintaining a moving average of the recent prediction error and the standard deviation of this error with a low pass filter rate of $0.1$. 
The raw surprise value is determined by calculating the current mean absolute difference between the event processing module's output $\mathbf{y}_{act}^t$ and the target value $\hat{\mathbf{y}}^t$. 
Thus, the larger the difference, the greater the surprise. 
We then pass this value through a suitably parameterized sigmoidal function, which yields zero for hardly any surprise and values close to one for values with large surprise.
The resulting surprise estimation value is then directly used as the input $x_s^t$ to the update gate.

\subsubsection{Event Boundary Anticipation}
Seeing that calculated surprise can only be processed after the fact, we next add an event boundary anticipation module, yielding SUGAR$_b$. 
This module aids the event processing module to preempt surprise and thus to generate fluid event transitions. 
It processes event boundary information as input and generates $x_s^t$ as output (Fig.~\ref{models}b).
While this module can indeed be trained to preempt surprise, our evaluations have shown that the learning progress is not as robust as desired and the developing latent event-predictive encodings are not fully stable. 
We thus continue with introducing counterfactual regularization to the event boundary anticipation module.

\subsubsection{Counterfactual Regularization (CFR)}

The event boundary anticipation module receives its gradient signal $\delta_{\zeta}^t$ from the event processing layer through the update gate.
We incorporate the counterfactual regularization as an additional term directly into the gradient equation of the update gate:
\begin{equation}
    \delta_{\zeta,reg}^t = \delta_{\zeta}^t + \beta \left(\sqrt{(y^t_{act} - \hat{y}^t)^2} - \sqrt{(y^t_{cf} - \hat{y}^t)^2}\right),
    \label{cfr}
\end{equation}
where $\beta$ indicates the gate status, taking on a value of 1 if the gate was open and 0 otherwise, although more nuanced and scaled activities of $\beta$ may be useful in the future for more complex systems.
$y_{act}^{t}$ is the actual prediction of the event processing network, while $y_{cf}^{t}$ is the alternative prediction had the gate been on the counterfactual state. $\hat{y}^{t}$ is the true label. 
During back-propagation $\delta_{\zeta,reg}^t$ thus regularizes the utility of gate openings. 

The comparison of the two possible prediction outcomes indicates the amount of error that was avoided or created, depending on whether the difference in Eq.~\ref{cfr} yields a negative or positive value, respectively. Note that a closed gate is unaffected by this additional loss component, since $\beta=0$.
CFR thus supports the opening of gates when the gate activity decreases prediction error (preventing surprise) by actively comparing it with what would have happened had the gate remained closed (cf. SUGAR$_c$, Fig.~\ref{models}c).
As a result of more selective gate openings, the latent event-predictive encodings $x_{o j}^{t}$ can be expected to develop more static latent codes, thus providing better guidance for generating the appropriate sequences via the event processing layer. 
In the future, this mechanism may be useful for gate control in more challenging applications, where overlapping event structures may need to be segmented by means of controlling an array of gates.

\section{Results}
We now evaluate the three SUGAR implementations, revealing its capabilities, justifying the addition of both the event boundary anticipation module and the CFR. 

\subsection{Calculated surprise}
\begin{figure}
\includegraphics[width=0.98\columnwidth]{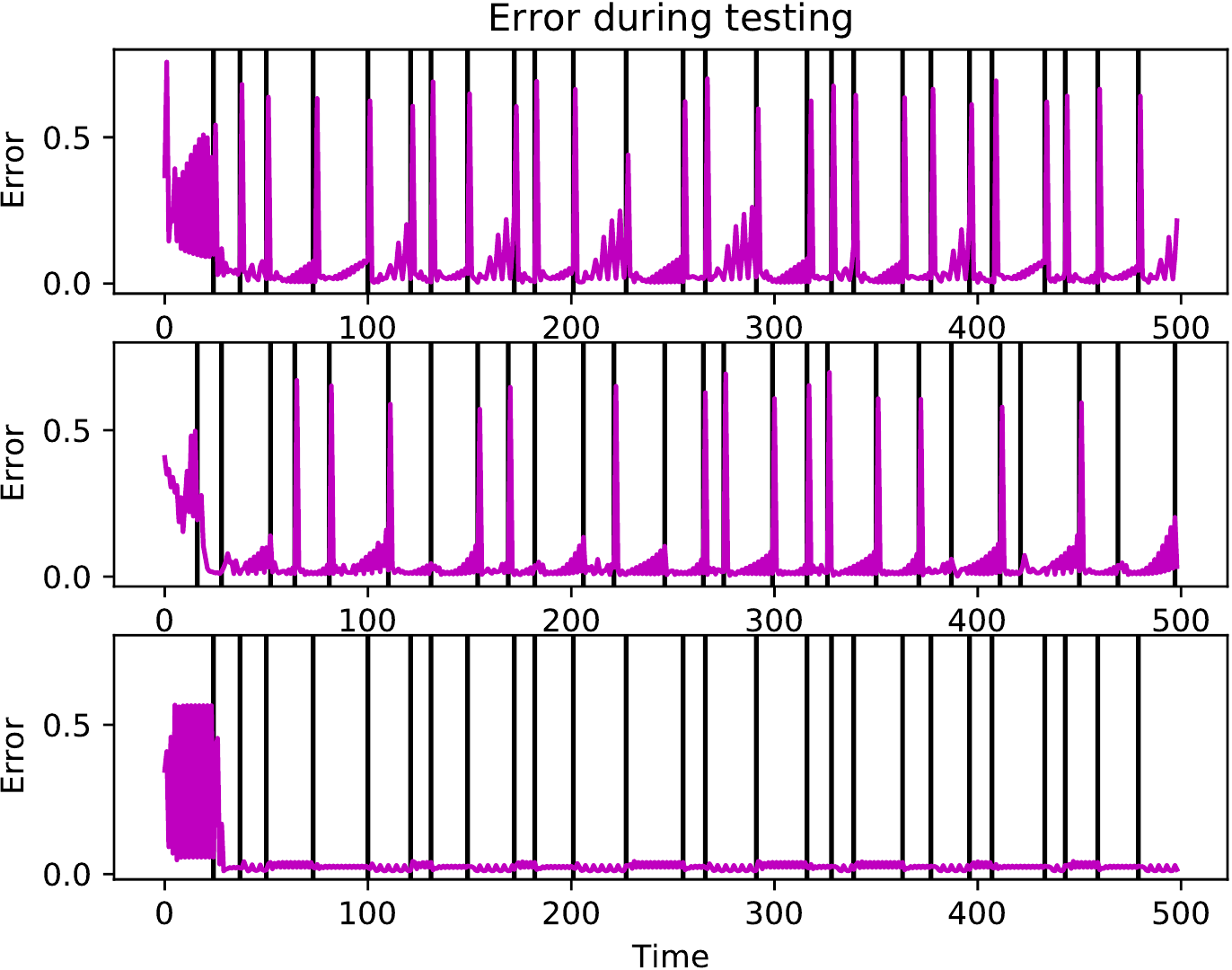}
\vspace{-.4cm}
\caption{Prediction error during testing when the update gate is always closed (top), controlled by SUGAR$_a$ (middle), or precisely opened only at event switches (fully informed SUGAR$_a$, bottom). Black lines indicate event transition boundaries. }
\label{predErrProGen}
\end{figure}

SUGAR$_a$ uses prediction error to calculate surprise.
This resulted in gate openings that occurred immediately after the switch between problems, in addition to several openings during the events. The resulting prediction error was larger when compared with the one obtained when the perfect surprise signal was provided, but smaller when the gate was kept closed (cf.~\autoref{predErrProGen}).
Thus, using calculated surprise signals to modulate top-down control improves performance. 

The error dynamics in \autoref{predErrProGen} (middle) show error spikes at the event transition boundaries, indicating that the event processing module indeed learns about the intra-event regularities in the data. 
However, even though surprise is registered, the reactive processing of surprise cannot prevent error spikes because surprise only offers post-hoc guidance, without any anticipatory prediction.
As shown in \autoref{predErrProGen} (bottom), providing ground-truth knowledge about upcoming event boundaries can clearly improve performance, effectively avoiding the error spikes. 
Thus, the performance of SUGAR can be significantly improved when the gate is able to open in anticipation of, rather than in reaction to surprising outcomes.

\begin{figure}[htb]
\includegraphics[width=0.9\columnwidth]{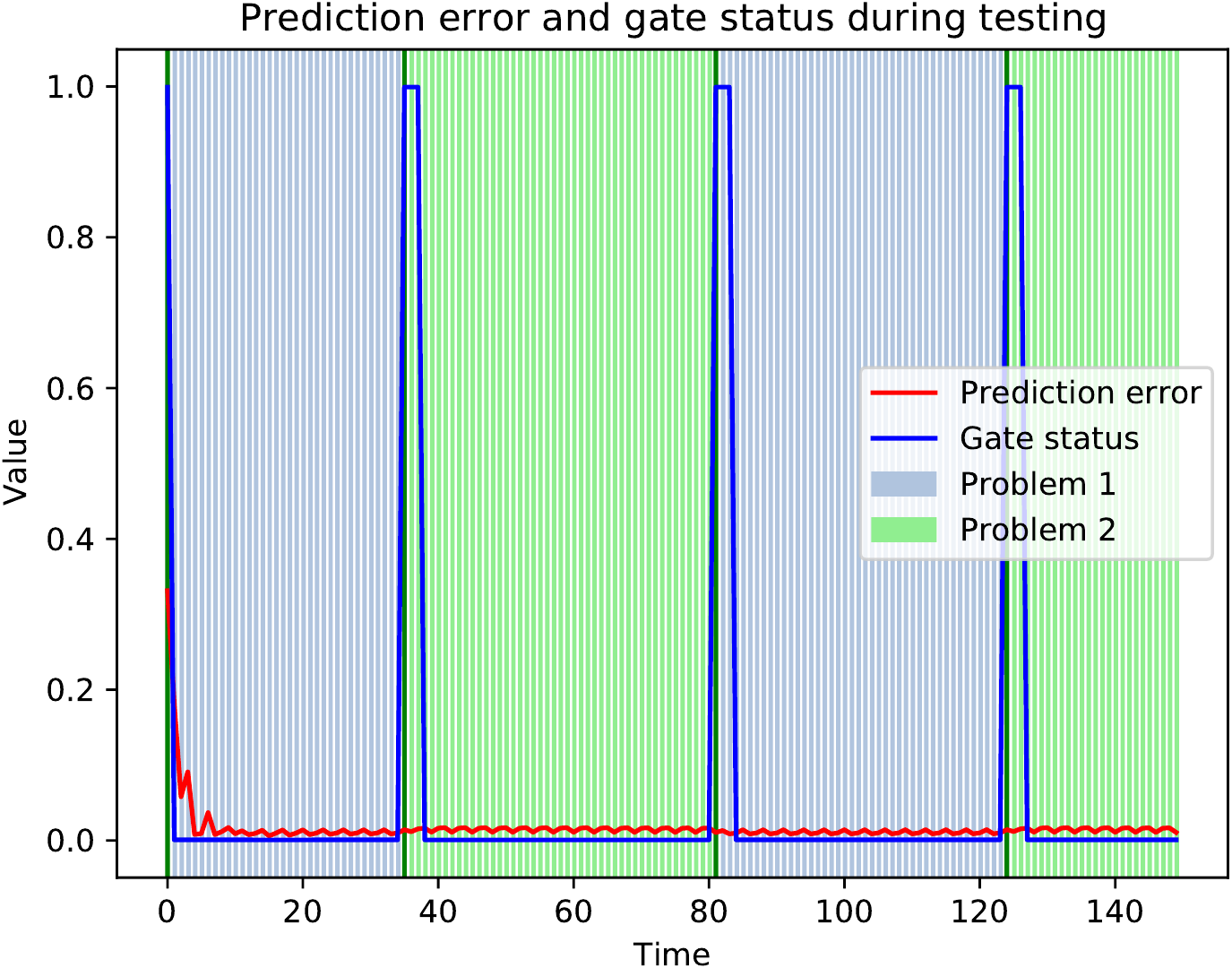}
\vspace{-.4cm}
\caption{Prediction error and surprise signals during testing on Problem1+2 using SUGAR$_b$ one hot plus event boundary indicator plus random input.}
\label{errSurpWithRandomInput}
\end{figure}

\subsection{Event Boundary Anticipation Module}
In order to achieve more fluid transitions between events, we extended the structure by adding the event boundary anticipation module as detailed above (SUGAR$_b$). 
This module receives event boundary information in the form of an increased value when a transition is about to happen. 
It utilizes this signal to control the gate, ideally opening it right before an event transition occurs, rather than right after.

We tested SUGAR$_b$ using Problem 1+2.
Using the event boundary anticipation module to modulate top-down control leads to predictions that are as accurate as the ones generated when the perfect surprise signal is simply provided just in time. Even when random binary values $[0,1]$ are added to the event boundary signal EB, the network still learns to recognize the ones that indicate the context switch and learns to open the gate at the right moment.
The prediction error dynamics shown in 
\autoref{errSurpWithRandomInput} indicate fluent transitions between events. The gate status additionally shows how gate spikes anticipate contextual switches.
This allows for signalling anticipatory, contextual encoding of events just before they switch, thus enabling more accurate predictions on the lower-level event processing layer.

\begin{figure}[t!]
\includegraphics[width=0.98\columnwidth]{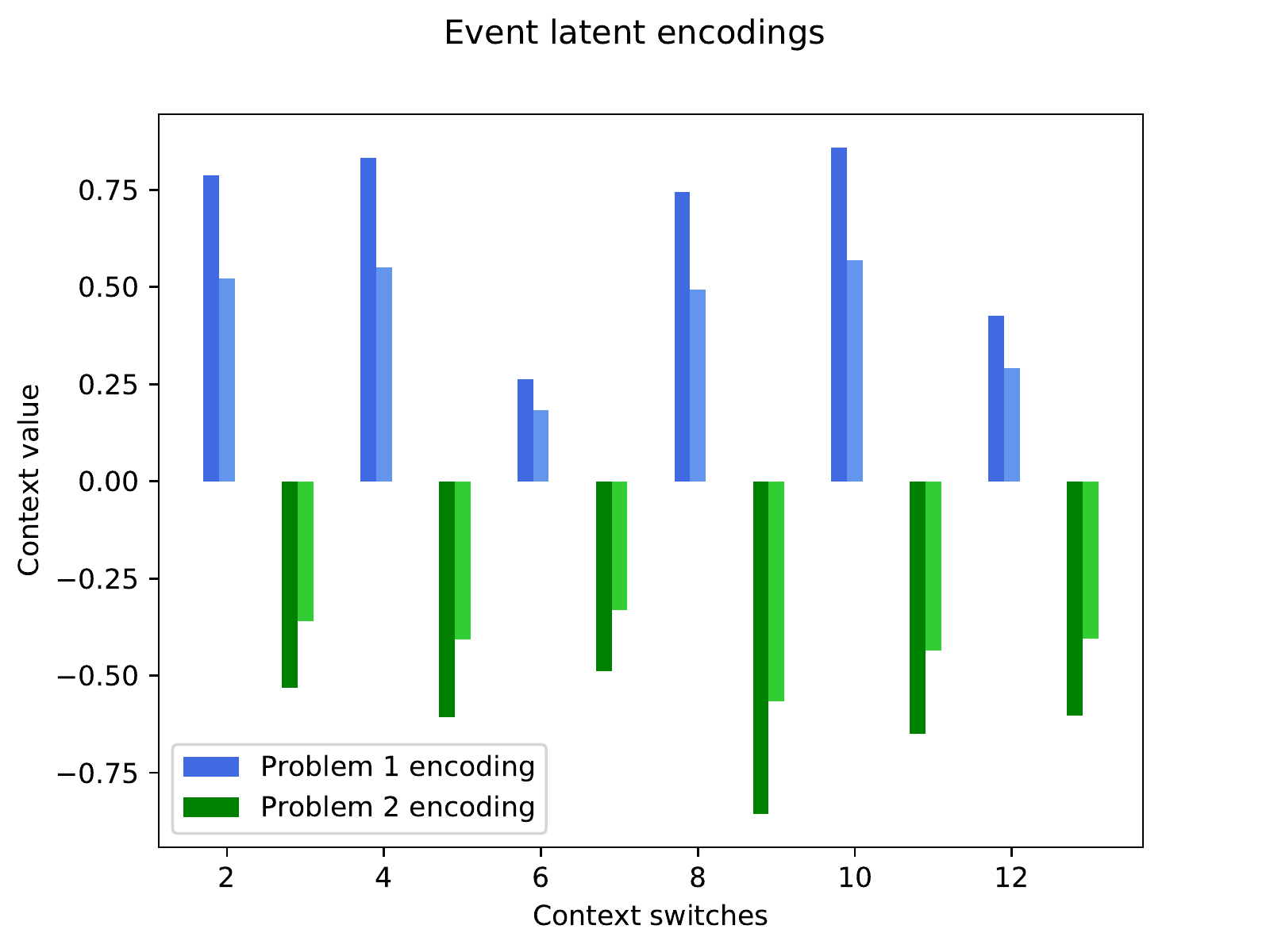}
\vspace{-.4cm}
\caption{Latent event encodings ($\mathbf{x}_{o}$) for Problem 1+2 emerging in SUGAR$_b$.} 

\label{eventEncodingsNoPenalty}
\end{figure}

\begin{figure}[t!]
\vspace{-.25cm}
\includegraphics[width=0.98\columnwidth,height=.6\columnwidth]{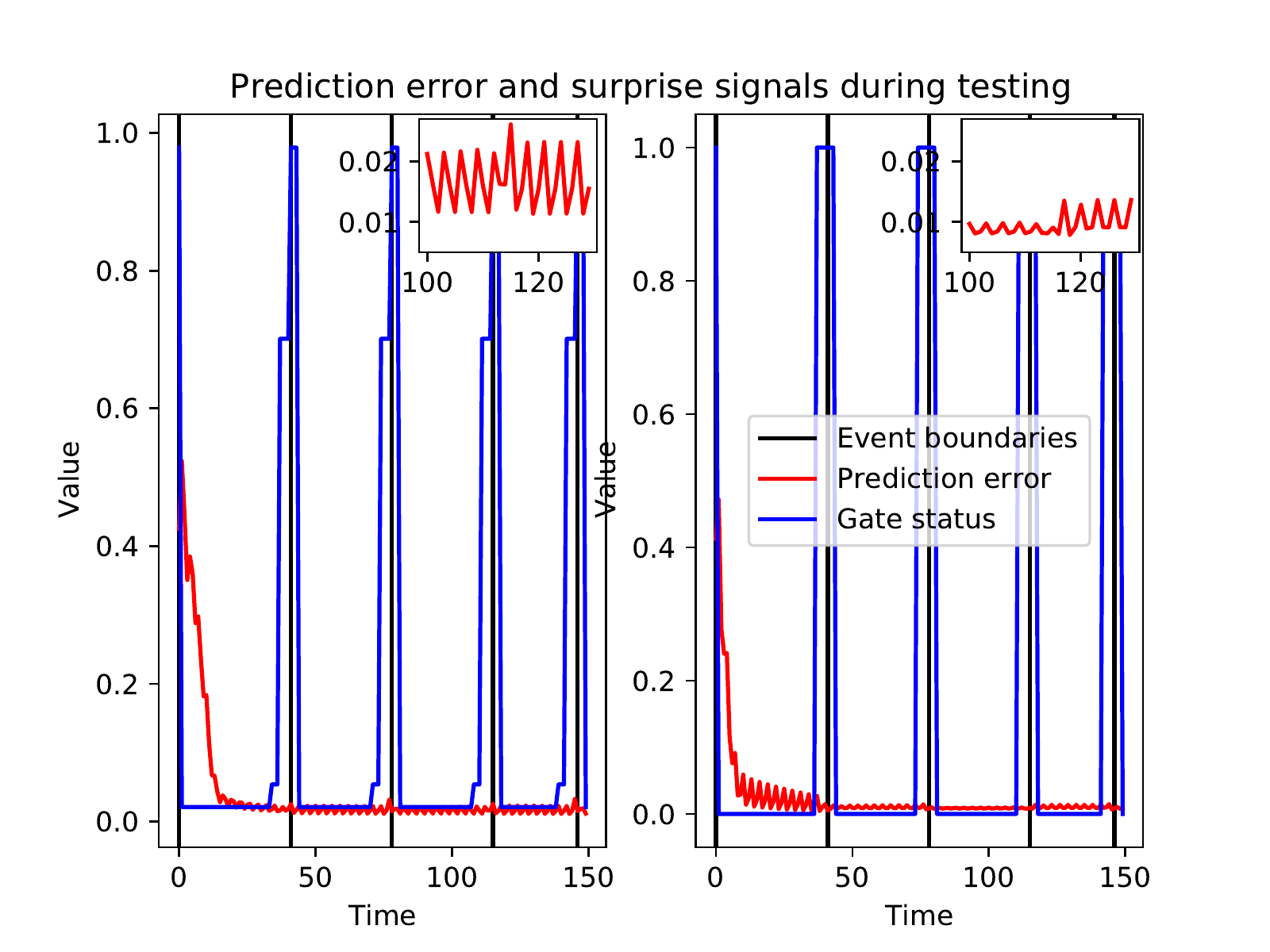}
\caption{Gate status and prediction error within events and at events boundaries for SUGAR$_b$ (left) and SUGAR$_c$ (right).
}
\vspace{-.4cm}
\label{surErrWithWithoutPenalty}
\end{figure}

\subsection{Counterfactual Regularization}
Surprise signals offer guidance about when the current event has ended and when to switch to the new context of the next event.
In the real world, we often receive signals from complementary cognitive systems that foreshadow when the current event is about to end (e.g. from the visual system as the perceived distance between one's hand and a glass of water approaches zero).
We mimicked this idea by gradually changing the input to the event boundary module before the switch, reaching a particular signal value just before the actual event switch.
The results in \autoref{surErrWithWithoutPenalty} confirm that this leads to gate openings in anticipation of the switch. 
However, closer inspections of the developing latent event codes revealed that these codes tended to dynamically change across subsequent event switches (Fig.~\ref{eventEncodingsNoPenalty}), which still results in slightly increased surprise signals between events (Fig.~\ref{surErrWithWithoutPenalty} left).

In order to increase stability during events, we thus added the CFR term (Fig.~\ref{models}c; Eq.~\ref{cfr}).
SUGAR$_c$ re-runs the event processing layer at all times steps $t$ when the gate was open with the counterfactual state (i.e. gate closed) of the event switching gate. 
Then, we compute the difference between the two prediction errors, thus inducing a penalty for non-beneficial gate openings (\autoref{cfr}).
As a result, the network avoids unnecessary gate openings.

\begin{figure}[t!]
\includegraphics[width=0.98\columnwidth]{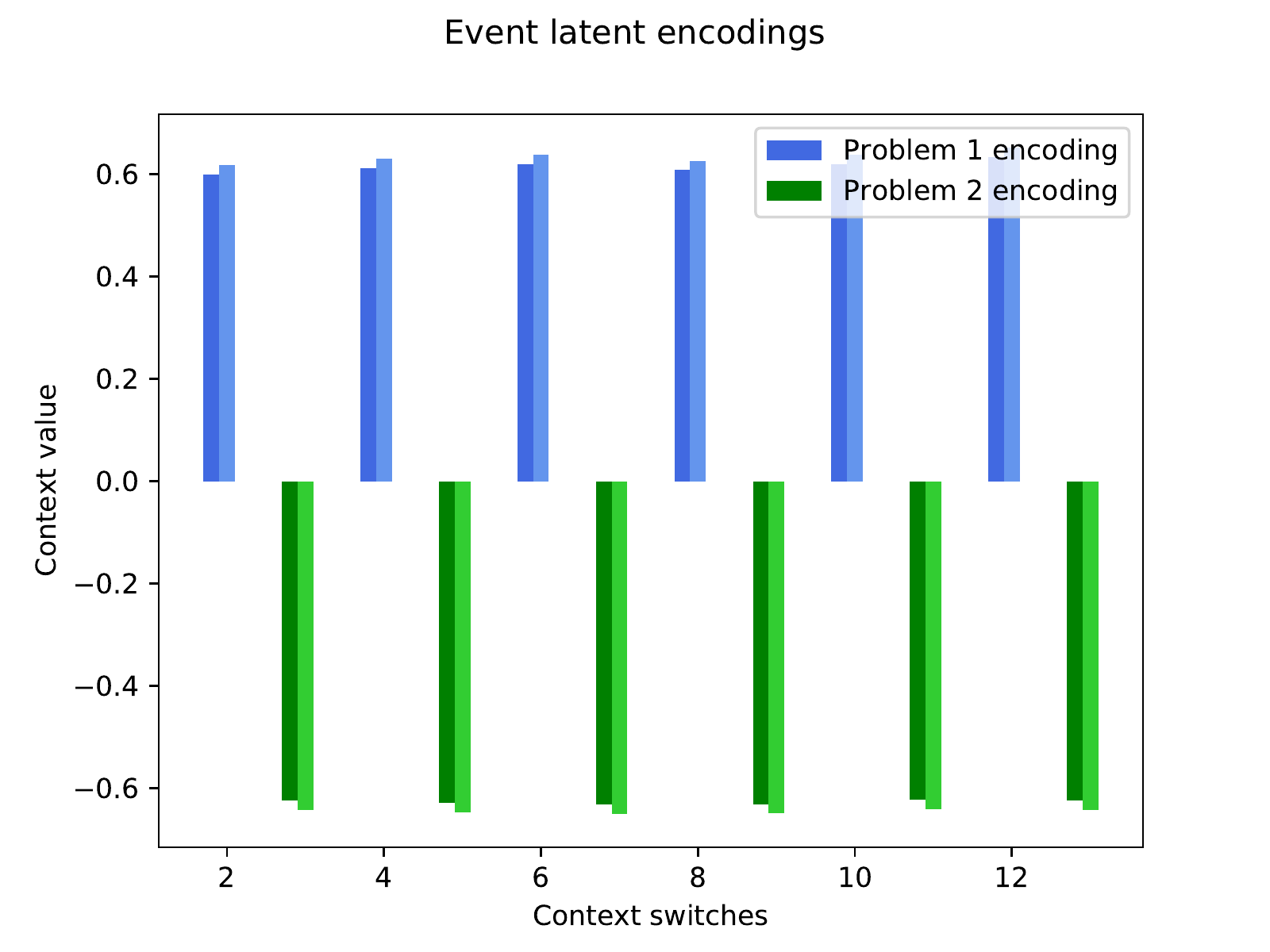}
\vspace{-.4cm}
\caption{Latent event encodings ($\mathbf{x}_{o}$) for Problem 1+2 emerging in SUGAR$_c$.} 
\label{eventEncodingsPenalty}
\end{figure}

Our results show that CFR indeed produces more stable latent event encodings, which do not change over time problem-respectively (cf. \autoref{eventEncodingsPenalty}).
Moreover, \autoref{surErrWithWithoutPenalty} shows that the consequent more focused gate openings in anticipation of an event switch fully avoid both error spikes and unnecessary gate activities during events (Fig.~\ref{surErrWithWithoutPenalty}, right-hand side).

\subsubsection{Latent event encodings}
Finally, we examine the latent event encodings $\mathbf{x}_o$ learned by the event anticipation layer.
Since CFR enforces maximally static, event-specific encodings, we expected that the top-down signals from the event switching layer $\mathbf{x}_o$ may capture similarities between the event-respective dynamics in the event processing layer.

Using Problem 1-3 (Fig.~\ref{comp_problem}), we investigated the latent event encodings developed in SUGAR$_c$ (Fig.~\ref{models}c) for the three problems.

\begin{figure}[t!]
\includegraphics[width=0.98\linewidth]{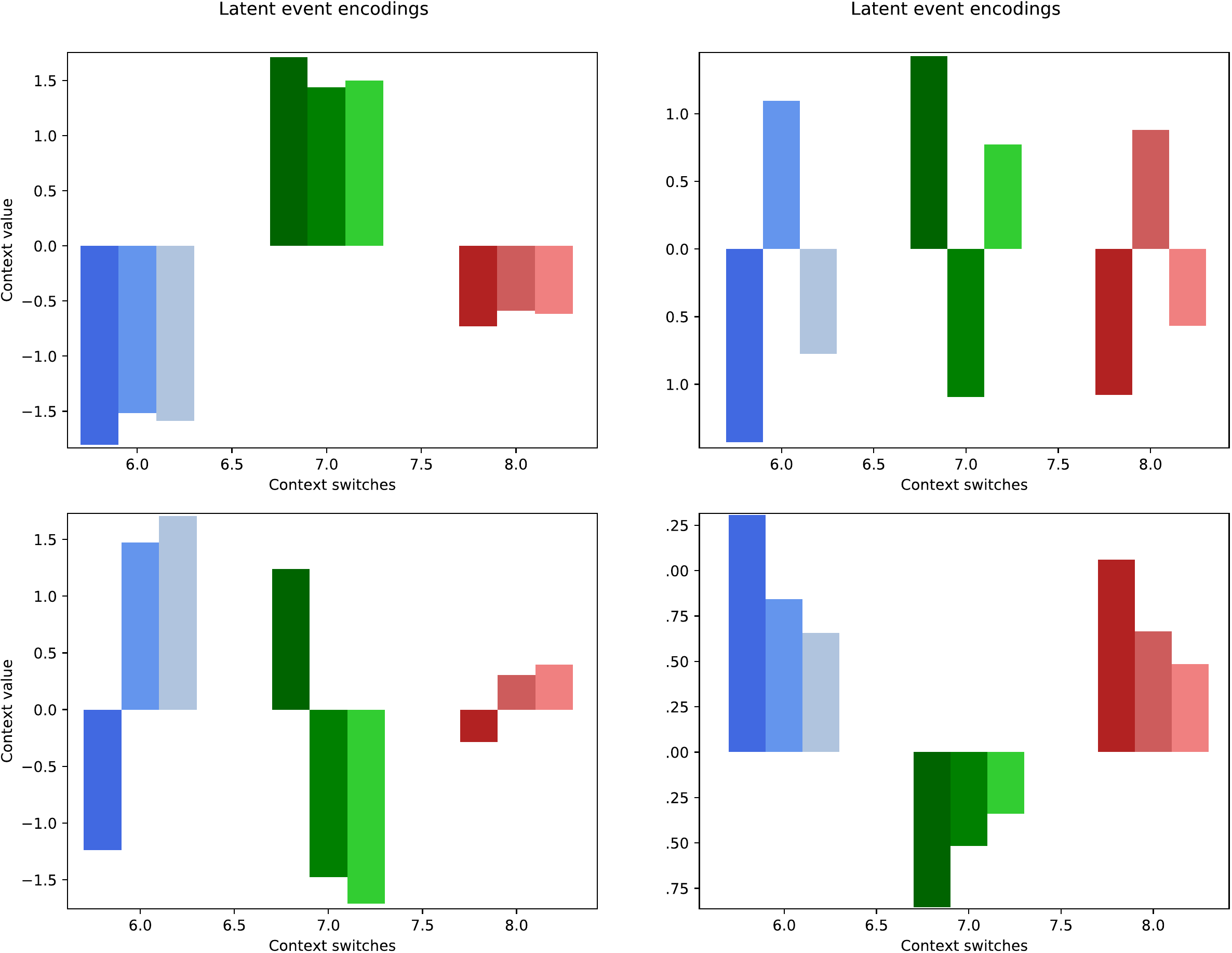}
\vspace{-.25cm}
\caption{Latent event encodings ($\mathbf{x}_{o}$) for four randomly initialized networks trained on Problem 1-3. In all cases, the code for Problem 3 (red bars) lies in between the code for Problem 1 and 2 (blue and green bars, respectively).} 
\vspace{-.25cm}
\label{comp_encodings}
\end{figure}

\autoref{comp_encodings} shows the respective encodings for different runs with differently initialized networks. Note that the encodings for Problem 3 are always more similar to that of Problem 1. In fact, they somewhat lie between the encodings of Problem 1 and Problem 3, but closer to Problem 1. This is because, apart from the fact that the symbols in Problem 1 are used in Problem 3 as well, Problem 3 starts with the same symbol sequence as does Problem 1. Still, there is a clear difference, since the symbol subsequence of Problem 2 is also contained in Problem 3.
Thus, the latent encodings tend to reflect properties of semantics and compositionality---in this case with respect to sequential structures---a claim that should certainly be investigated further.

\section{Discussion}
In this work, we introduced SUGAR---a normative neural network model that yields event-predictive encodings, which exhibit properties suggestive of compositionality. These encodings provide top-down contextual guidance for predicting hierarchically generated sequences, where a gating layer modulates this control. We compared both prediction error trained surprise as well as using noisy signals about event boundaries. For the latter to be effective, we introduce a novel counterfactual regularization term, which yields more accurate predictions, particularly at event boundaries, and more stable latent event encodings. 

An important limitation in our current work is the need to provide an external signal that foreshadows event switches. This was inspired by the fact that orienting reflexes or other sensory signals often provide good indications for impending event boundaries \cite<e.g., when our hand touches a glass or when we reach the beginning of a flight of stairs;>{Baldwin2021,Kuperberg2021}. Future work could improve on these methods by using anticipatory signals from the event-predictive processing module to prepare for event transition processing, possibly in an attention-based manner.
Additionally, the scalability of the network as well as other types of event-like sequence processing tasks \cite<e.g.,>{Humaidan2020} should be investigated.
We expect to thus enable the networks to guide event processing even more effectively by means of the developing, predictive process-focusing, top-down event encodings.

\section{Conclusion}
Event cognition plays a crucial role in providing contextual guidance for solving problems and interpreting the world around us \cite{Radvansky2014}. An understanding of the correct event context is important to foster the effective compression of the sensorimotor processing stream into compositional latent encodings for both artificial and biological agents \cite{gershman2010learning}. Yet in order to effectively learn hidden structures, inductive learning biases are necessary \cite{Battaglia:2018,franklin2020structured}. Here we studied how the gating of top-down control via surprise and the use of counterfactual regularization can help for both avoiding unnecessary gate openings and learning more stable, latent event-predictive encodings with compositional properties. 
Seeing that event-predictive structures are closely interlinked with language structures and can suitably simulate explanatory thought processes \cite{Butz:2021}, we believe that SUGAR-based architectures with counterfactual regularization mechanisms may be useful for developing useful hierarchical and compositional encodings in other domains and tasks---and thus particularly for developing more advanced machine learning systems \cite{Butz:2021a}.
Our results provide preliminary evidence that gating mechanisms combined with counterfactual error regularization are an important inductive learning bias, which may help bridge the gap between artificial and human intelligence.

\section{Acknowledgments}
This research was funded by the German Research Foundation (DFG) within Priority-Program SPP 2134 - project ``Development of the agentive self'' (BU 1335/11-1, EL 253/8-1) and under Germany’s Excellence Strategy – EXC 2064/1 – 390727645.
The authors thank the International Max Planck Research School for Intelligent Systems (IMPRS-IS) for supporting DH and CG and the German Federal Ministry of Education and Research (BMBF): Tübingen AI Center, FKZ: 01IS18039A for supporting CMW.

\bibliographystyle{apacite}

\setlength{\bibleftmargin}{.125in}
\setlength{\bibindent}{-\bibleftmargin}

\bibliography{CogSci_Template}

\end{document}